%% file: main.tex
\documentclass[lettersize,journal]{IEEEtran}

\newif\ifarxiv
\arxivtrue %

\usepackage{array}
\usepackage[caption=false,font=normalsize,labelfont=sf,textfont=sf]{subfig}
\usepackage{stfloats}
\usepackage{graphicx}
\usepackage[dvipsnames]{xcolor}
\usepackage{soul}
\usepackage{threeparttable} %
\usepackage{hyperref}
\usepackage{adjustbox}
\usepackage{threeparttable}
\usepackage{booktabs}
\usepackage{multirow}
\usepackage{siunitx} %
\usepackage{amssymb} %
\usepackage{amsmath,amsfonts} %
\usepackage{bm} %
\usepackage{pifont} %
\usepackage{cleveref} %
\usepackage{microtype} %
\usepackage[
  backend=biber,
  style=ieee,
  maxnames=999,
  maxbibnames=999
]{biblatex}

\newcommand{\abbr}{ECL}

\newcommand{\cmark}{\ding{51}}%
\newcommand{\xmark}{\ding{55}}%

\usepackage{epsf,caption,subcaption}
\usepackage{amsmath}

\makeatletter
\g@addto@macro\@eqnnum{\aftergroup\noindent}
\makeatother

\newcommand{\Figref}[1]{\hyperref[#1]{Figure~\ref{#1}}}
\newcommand{\figref}[1]{\hyperref[#1]{Fig.~\ref{#1}}}

\AtNextBibliography{\footnotesize} %
\AtEveryBibitem{\clearfield{doi}} %
\AtEveryBibitem{\clearfield{url}\clearfield{eprintclass}} %
\AtEveryBibitem{\clearfield{urldate}}
\AtEveryBibitem{\clearfield{editor}} %
\DeclareFieldFormat[misc]{title}{``#1''} %
\DeclareFieldFormat{eprint:arxiv}{\mkbibemph{arXiv:#1}} %
\DeclareFieldFormat[online]{title}{\mkbibemph{#1}} %
\usepackage[font=footnotesize,labelsep=period]{caption}

\begin{document}

\title{Versatile On-device Adaptation at the Edge\\ by Unifying Few-shot, Zero-shot, Continual, and In-context Learning}

\author{Douwe den Blanken,
\IEEEmembership{Graduate Student Member, IEEE},
Martin Lefebvre,
\IEEEmembership{Member, IEEE}
\\
and
Charlotte Frenkel,
\IEEEmembership{Member, IEEE}

\vspace{-2em}

\thanks{%
\ifarxiv
This work has been submitted to the IEEE for possible publication. Copyright may be transferred without notice, after which this version may no longer be accessible.

\fi
This publication was funded by the Dutch Research Council (NWO) as part of the projects \textit{AdaptEdge} (file number 20267 in the \textit{NWO Talent Programme -- Veni}) and \textit{Transforming the Adaptability of Decentralized AI} (file number NGF.1609.242.038 in the \textit{NGF - AiNed AiNed XS Europe} programme).

Douwe den Blanken, Martin Lefebvre and Charlotte Frenkel are with the Microelectronics Department (EEMCS Faculty), Delft University of Technology, 2628 CD Delft, Netherlands (e-mail: d.m.j.denblanken@tudelft.nl; m.lefebvre@tudelft.nl; c.frenkel@tudelft.nl; \textit{Corresponding author: Douwe den Blanken}).}%
}

\ifarxiv
    \markboth{}{}
\else
    \markboth{IEEE Transactions on Circuits and Systems for Artificial Intelligence}{}
\fi

\emergencystretch 3em %

\newcommand{\eg}{\textit{e.g.},~}
\newcommand{\etal}{\textit{et~al.}}

\newcommand{\omnfwayoshot}{96.8 ± 1.6\%} %
\newcommand{\omnfwayfshot}{98.8 ± 0.5\%} %
\newcommand{\omntwayoshot}{89.1 ± 1.3\%} %
\newcommand{\omntwayfshot}{96.1 ± 0.5\%} %
\newcommand{\omnthwayoshot}{83.3 ± 1.2\%} %

\newcommand{\red}[1]{\textcolor{red}{#1}}

\maketitle

\begin{abstract}
With the ever-increasing pervasiveness of smart edge devices, the demand is growing for applications that can be tailored to users (\textit{e.g.},~custom keyword spotting) or patients (\textit{e.g.},~adaptive health monitoring). Yet, most edge devices rely on fixed inference algorithms and thus cannot learn on-device to personalize predictions.
When they can, devices typically support only a specific learning scenario, such as few-shot learning (FSL):
going beyond this requires resorting either to another specialized device or to cloud-based retraining, which implies significant energy and latency overheads, a lack of real-time capabilities, and privacy concerns.
In this work, we introduce embedder-centric learning (\abbr), a framework that unifies four different online learning scenarios:
FSL for on-the-fly customization, continual learning (CL) for knowledge accumulation, zero-shot learning (ZSL) for leveraging semantic data, and in-context learning (ICL) for adapting beyond classification. 
We demonstrate in silicon that \abbr~can be deployed on resource-constrained devices across four real-world use cases representative of the aforementioned learning scenarios.
Our approach establishes a new state-of-the-art performance for FSL character recognition (Omniglot: 96.8\% for 5-way 1-shot, 83.3\% for 32-way 1-shot), and the first hardware baseline for CL in keyword spotting (NeuroBench keyword FSCIL: 71.8\% for 200-way 5-shot). Moreover, we present the first hardware demonstrations of ZSL with semantic data (60.6\% for 5-way spoken sentence classification) and ICL (46.2\% at the 500th token of RegBench) operating at micro-to-milliwatt power budgets.
Therefore, by unifying multiple learning scenarios, we pave the way for smart and versatile devices that can adapt right at the edge, without reliance on the cloud.
\end{abstract}

\section{Introduction}

\IEEEPARstart{I}{n} the last decade, the integration of neural networks (NNs) on resource-constrained devices at the edge has become increasingly commonplace.
Example use cases include navigation for drones~\cite{dronesurvey}, keyword spotting (KWS) on smart speakers~\cite{vocell,kwantae_kim_23_uw_ring_oscillator_ro_kws,tan20241_8khz,park_rebuttal_kws}, and health monitoring on wearables~\cite{liu2023ultra,wearable_cardiac_monitoring_system,smart_ecg,clockless_ecg,eeg_ecg_combined_chip_liu}.
However, most of these devices rely on fixed, pre-trained NNs that cannot be adapted post-deployment.
Consequently, these devices cannot locally handle shifting data distributions, emerging features, specific or additional users, or evolving task requirements~\cite{reckon_frenkel},
such as the sim-to-real gap for drones, new keywords for KWS, and user- or patient-specific tailoring for wearables.

\begin{figure}[t]
\centering
\includegraphics[width=0.45\textwidth]{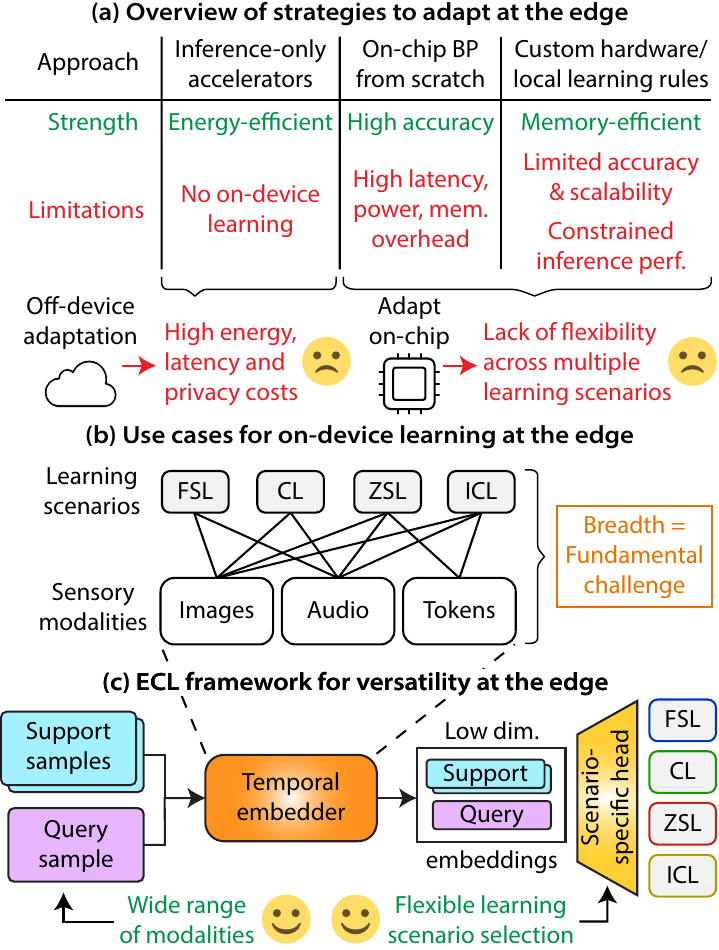}
\caption{\textbf{(a)} Overview of current approaches to learning at the edge and their limitations. \textbf{(b)} Learning at the edge requires supporting different learning scenarios, such as few-shot learning (FSL), continual learning (CL), zero-shot learning (ZSL), and in-context learning (ICL), while accommodating for different sensory modalities, forming a fundamental challenge. \textbf{(c)} Outline of the proposed embedder-centric learning (ECL) framework, which splits each learning scenario into a shared embedder and a scenario-specific head. This embedder-centric framing lets \abbr~support a wide range of sensory modalities while unifying four learning scenarios.}
\label{fig:fig0}
\end{figure}

Addressing these needs is challenging, due to the various limitations of existing strategies to adapt (\autoref{fig:fig0}(a)).
We group these strategies into three categories.
The first category, comprising inference accelerators, relies on the cloud for \mbox{off-device} adaptation, which comes at the expense of the energy cost for the cloud link~\cite{jain2023tinyvers,verhelst2017embedded_privacy_and_latency_tx_energy}, latency penalties that preclude online learning~\cite{sze2017efficient_privacy_and_latency, verhelst2017embedded_privacy_and_latency_tx_energy}, and risks of exposing user private data~\cite{sze2017efficient_privacy_and_latency,verhelst2017embedded_privacy_and_latency_tx_energy}.
The second trains a model from scratch with backpropagation (BP) directly on the device~\cite{onchip_sgd_from_scratch_kim2015640m, onchip_sgd_from_scratch_gonugondla201842pj,onchip_sgd_from_scratch_amravati201855nm} but requires storing all intermediate activations, which is especially prohibitive for long temporal signals~\cite{reckon_frenkel}.
The third aims to alleviate this overhead of BP by implementing specialized learning algorithms in hardware~\cite{fslhdnn,clo_hdnn,sapiens,other_fsl_cim}, or by exploring algorithms based on local learning rules~\cite{stdp_local_learning_chen20184096, reckon_frenkel}.
However, while multi-layer variants start emerging~\cite{tess_marco, ostl_bohnstingl}, their hardware implementation remains an open challenge to the best of our knowledge. Furthermore, custom learning-optimized hardware typically trades inference efficiency for learning accuracy~\cite{blanken2025chameleonmatmulfreetemporalconvolutional}. %
Therefore, enabling on-chip learning at minimal hardware cost remains a challenge.

Beyond efficiently supporting a \textit{single} learning scenario on-chip, real-world deployment must cope with diverse data dynamics, label availability, and task structures. This demands support for additional learning scenarios, also across sensory modalities, from images to long-timescale audio (\autoref{fig:fig0}(b)).
This breadth forms a fundamental challenge: %
current accelerators cannot accommodate more than one learning scenario, yet designing a specialized chip for every combination of scenario and sensory modality is impractical.
Therefore, we aim to answer the following question: \textit{How can we optimally support different learning scenarios for versatility and efficiency across use cases at the edge?}

\looseness-100  In this work, we propose the embedder-centric learning (\abbr) framework, which unifies few-shot learning (FSL), continual learning (CL), zero-shot learning (ZSL), and in-context learning (ICL) with support for a wide range of sensory modalities (\autoref{fig:fig0}(c)).
Our core idea is to frame each learning scenario in an embedder-centric way, splitting it into two components: a strong embedder NN, and a scenario-specific head of one or more fully-connected (FC) layers that processes the resulting \textit{embeddings} into a prediction.
This embedder-centric design is effective for two reasons.
First, embeddings represent data in a reduced-dimensional space, keeping the knowledge memory small enough to maintain and reuse on-device across FSL, CL, ZSL, and ICL.
Second, an embedder that excels on \textit{temporal} data, especially across long temporal dependencies, lets \abbr~support different sensory modalities efficiently: audio (provided raw or sequentially as Mel-frequency cepstral coefficients (MFCCs)~\cite{davis1980comparison_mfcc} frames), images (provided sequentially as pixels), or, more generally, tokens.

\looseness-100  We validate the performance and low cost of the \abbr~framework for each of the learning scenarios both in software and hardware using our recently introduced Chameleon system-on-chip (SoC)~\cite{blanken2025chameleonmatmulfreetemporalconvolutional}.
We surpass state-of-the-art (SotA) FSL accuracies on the Omniglot dataset~\cite{lake_omniglot} (96.8\% for 5-way 1-shot, 83.3\% for 32-way 1-shot),
and we demonstrate for the first time on-device CL on the NeuroBench keyword few-shot class-incremental learning (FSCIL) dataset to classify 200 classes at a \qty{9.5}{\micro\watt} real-time power.
We also perform ZSL using semantic data for the first time on-device, using only \qty{3.1}{\micro\joule} to learn five new classes.
In addition, we present the first demonstration of on-device ICL on a formal language, requiring only \qty{16.8}{\micro\joule} per token.
These results establish new baselines across learning scenarios and sensory modalities, and underline our embedder-centric approach as an enabler for learning at the edge.

\begin{figure}[t]
\centering
\includegraphics[width=0.43\textwidth]{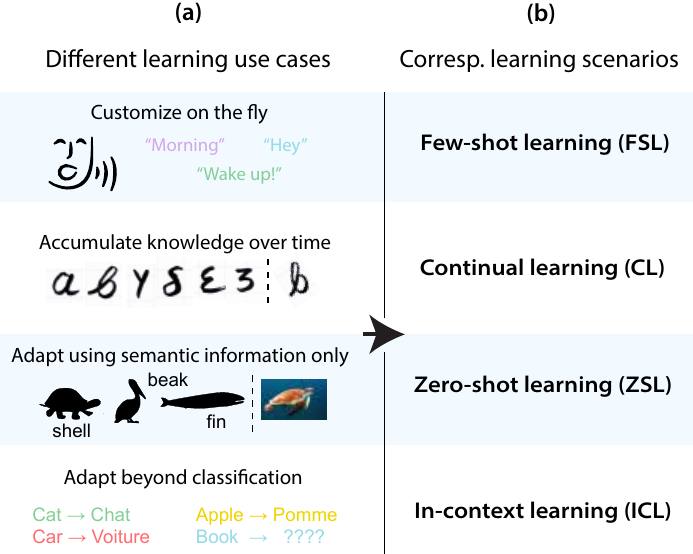}
\caption{\textbf{(a)} Comparison of four different learning use cases, from customizing to new tasks on the fly to adapting beyond classification tasks. \textbf{(b)} To enable each of these use cases, a learning scenario can be adopted: FSL, CL, ZSL, or ICL, each corresponding to a use case.}
\label{fig:iia}
\end{figure}

\begin{figure*}[t]
\centering
\includegraphics[width=0.77\textwidth]{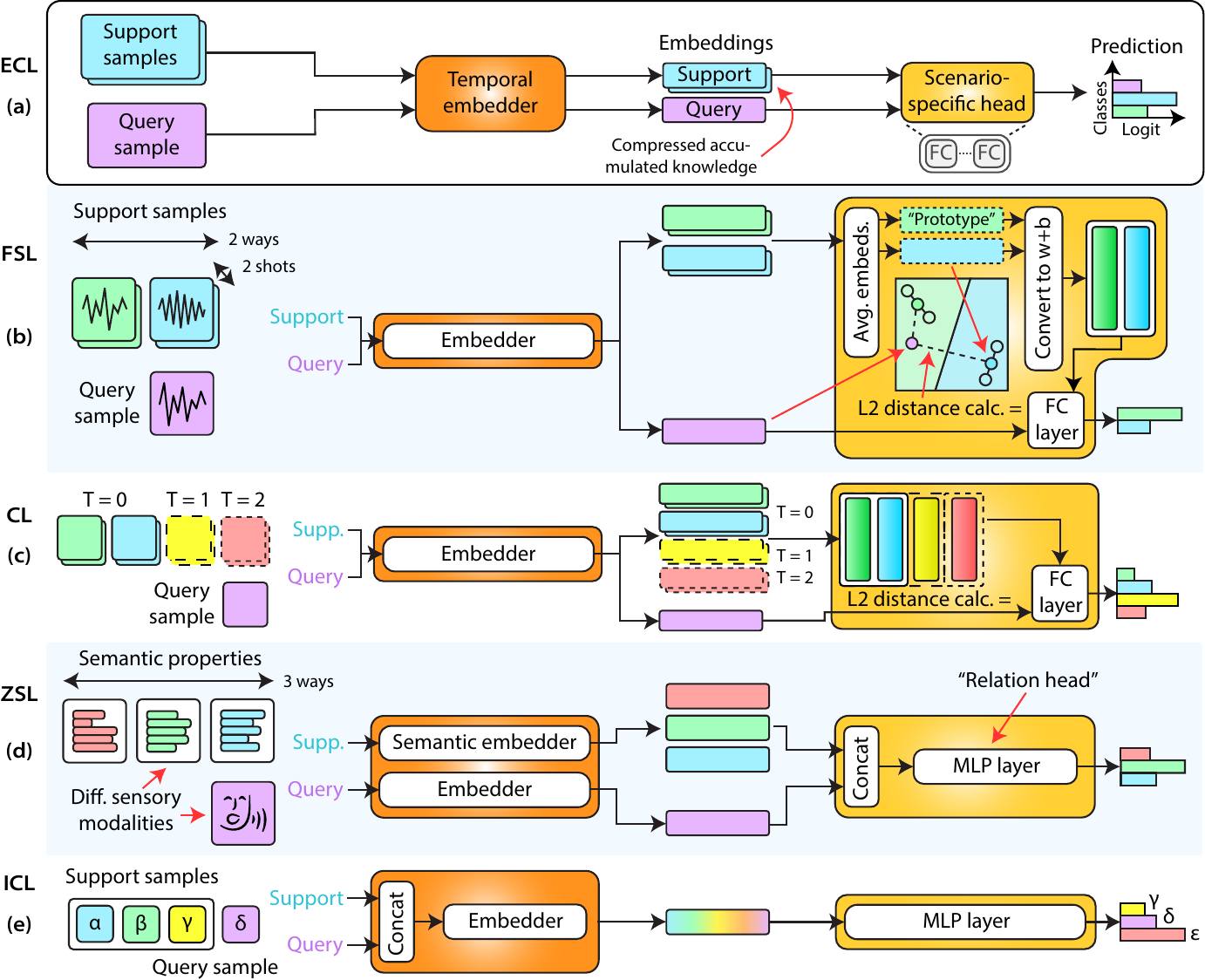}
\caption{Overview of the \abbr~framework. \textbf{(a)} \abbr~unifies four online learning scenarios through an embedder-centric formulation: support and query samples are embedded before a scenario-specific head, consisting of one or more FC layers, produces a prediction on the query sample. \textbf{(b)} FSL in \abbr~uses prototypical networks (PNs)~\cite{snell2017prototypical}: support embeddings are averaged class-wise into prototypes, and the query sample is assigned to its nearest prototype.
Prototypes are transformed into an equivalent representation in terms of weights and biases (w + b) of an FC layer, to compute the distance between the query sample and the prototypes efficiently.
\textbf{(c)} CL in \abbr~reuses the same PN-based approach, but appends
the parameters corresponding to a new class (learned at time $T\geq1$) to existing FC layers' parameters (learned during an initial training phase at $T=0$), instead of overwriting them.
\textbf{(d)} ZSL in \abbr~uses a relation network (RN)~\cite{sung2018learning} with two embedders, one for the semantic support data and one for the query sample; their embeddings are concatenated and passed to a relation head, here an multi-layer perceptron~(MLP), for the final prediction. \textbf{(e)} ICL in \abbr~concatenates the query and support samples before embedding, and the scenario-specific head feeds the joint embedding to an MLP to perform the prediction.}
\label{fig:fig1}
\end{figure*}

The remainder of this article is structured as follows. First, \autoref{sec:use_cases} introduces the four learning use cases at the edge and the corresponding learning scenarios considered in this work.
Then, \autoref{sec:ECL} presents our proposed \abbr~framework and details its implementations of the FSL, CL, ZSL, and ICL learning scenarios. Then, \autoref{sec:results} outlines our test setup and presents hardware and software results for each learning scenario, followed by a synopsis of the overall results. Finally, we offer concluding remarks in \autoref{sec:conclusion}. To promote reproducibility, reuse, and improvement, all code for this paper is open-source, including the training frameworks, accelerator source code, and the test/simulation setup.\footnote{\url{https://github.com/cogsys-tudelft/ecl}}

\section{Background: On-device Learning Use Cases at the Edge}
\label{sec:use_cases}

Figs.~\ref{fig:iia}(a) and (b) respectively depict four different use cases for learning at the edge and the learning scenarios that are used to tackle them.
The first highlighted use case is to customize to new data or tasks on the fly.
Examples of this include learning with only a few examples to detect new keywords~\cite{mazumder2021few}, to classify novel image classes~\cite{vinyals2016matching}, and to detect custom gestures~\cite{pfister2014gesture}.
\textit{Few-shot learning (FSL)} addresses this problem of recognizing new classes that were not seen during pre-training using only a few labeled samples per class. 
The new classes are referred to as \textit{ways} while the samples per class are referred to as \textit{shots}~\cite{maml_finn, snell2017prototypical, lake2011one, koch2015siamese, fei2006one}.
As FSL only requires limited data, it is a particularly good fit for the edge, where the online appearance of new data is often scarce (\textit{e.g.},~examples given by users or rare events not accommodated for during training). FSL, however, requires every new class to be available at once.

When classes instead emerge incrementally, the model must accumulate knowledge over time rather than acquire it in a single step.
Examples include incrementally learning additional image classes~\cite{tao2020few_fscil}, such as a growing alphabet of handwritten characters, or learning to detect an expanding set of keywords~\cite{yik2025neurobench}.
To handle this second set of use cases, \textit{continual learning (CL)} can be applied.
CL allows an NN to gain knowledge of new tasks or data distributions over time, without forgetting the previously acquired knowledge~\cite{kirkpatrick2017overcoming}. However, both FSL and CL still depend on labeled data, which the edge cannot always provide.

A third category of use cases instead offers only semantic information from different sensory modalities, such as high-level attributes, audio, or textual descriptions.
Examples of this include animal classification from high-level properties only~\cite{lampert2013attribute} or spoken-sentence detection from transcriptions alone.
\textit{Zero-shot learning (ZSL)} enables learning from such semantic information alone, removing the need for labeled test-domain data for each new task~\cite{lampert2013attribute, snell2017prototypical}. In effect, ZSL learns new tasks using a sensory modality different from that of the test data.

Each scenario so far stops at classification. Yet many edge use cases call for more, for example, forecasting temporal data or few-shot language modeling, forming the fourth and final group of use cases.
Using \textit{in-context learning (ICL)}, use cases such as regression or next-token prediction tasks~\cite{JMLR:v25:23-1042incontextregression, garg2022can_incontextregression, brown2020language}, where the next symbol should be predicted given a vocabulary and labeled input-output pairs, can also be dealt with.
ICL relies purely on the forward pass of a model: learning happens by building memory associations between a sequence of labeled examples combined with a test input inside a context window, without modifying any model parameters~\cite{brown2020language}.  
ICL was first observed as an emergent property of the transformer architecture~\cite{vaswani2017attention} in large-scale language modeling~\cite{brown2020language}. However, it has recently been shown that it is not exclusive to these models~\cite{tong2024mlps}.

Supporting these four learning scenarios at the edge conventionally requires a separate specialized system for each. However, this is impractical under the tight compute and memory budgets of edge devices. Hence, how to efficiently support all four within one device remains an open problem.

\section{The \abbr~Framework for Versatility at the Edge}\label{sec:ECL}

In this work, we propose a framework called \abbr, that unifies the above four learning scenarios by framing each of them in an embedder-centric way. \abbr~thus allows splitting each learning scenario into two key components illustrated in \autoref{fig:fig1}(a): (i) an embedder NN and (ii) a scenario-specific head.
First, the NN maps support samples, the labeled data for learning, and a query sample, the input to process after learning, to an \textit{embedding}, \textit{i.e.}, a feature vector of substantially lower dimensionality than the input.
Together, the embedded support samples represent a compressed form of accumulated knowledge.
Second, the scenario-specific head processes embedded knowledge together with the query embedding using one or several fully-connected (FC) layers, in a way that is dependent on the learning scenarios.
Through this shared structure, \abbr~efficiently supports versatile adaptation.

While our framework removes the need for designing a specialized implementation per scenario, supporting adaptation across sensory modalities, however, would still require a different NN type for each.
To avoid this, \abbr~requires an NN that can embed samples containing long temporal dependencies, such as audio consisting of MFCC frames or even raw samples, images converted into pixel streams, or effectively any data type provided as tokens.
We therefore propose to use temporal convolutional networks~(TCNs)~\cite{bai2018empirical} as the temporal embedder for \abbr.
TCNs are NNs that use stacked causal 1D convolutions combined with residuals to capture long-range relationships in sequential data~\cite{bai2018empirical}.
These NNs can extract high-quality embeddings from long sequences while their memory cost scales only logarithmically with sequence length~\cite{blanken2025chameleonmatmulfreetemporalconvolutional}, allowing them to efficiently support a wide range of sensory modalities.
While a new embedder still has to be pre-trained off-chip for each new sensory modality, the target deployment platform for \abbr~only needs to support one NN type, simplifying the hardware requirements.

Independent of the chosen embedder, by construction, the embedder-centric stance of \abbr~itself yields three further advantages for hardware implementation.
First, the target hardware only needs to support regular NN inference, since generating an embedding is equivalent to a forward pass. Second, the knowledge memory that stores embeddings can stay small (a few \qty{}{\kilo\byte}), because embeddings have far lower dimensionality than the input samples. Third, formulating the scenario-specific head as one or more FC layers lets \abbr~execute within the same NN inference pipeline, requiring only minor control logic to construct the layer. Overall, \abbr~avoids learning-specific memories and introduces only minor control overhead while relying on standard inference hardware, bringing efficient learning across FSL, CL, ZSL, and ICL to extreme-edge devices.

\vspace{0.7em}

Furthermore, let us now explain how the \abbr~framework implements each learning scenario~(\autoref{fig:fig1}), by detailing the operation of the embedders and scenario-specific heads.

\subsubsection{FSL Implementation in \abbr}

For FSL deployment, we use a technique called prototypical networks (PNs)~\cite{snell2017prototypical}.
In PNs, an embedder NN is used to embed the available support samples. A \textit{prototype} for a new class is then formed by averaging the embeddings from the support samples for that class (\autoref{fig:fig1}(b)).
After embedding, an unknown query sample is then classified as the class of the prototype with the lowest L2 distance to its embedding~\cite{snell2017prototypical}.
To efficiently support PNs in \abbr, we employ the equivalent transformation of PNs into weights and biases that parametrize a single FC layer~\cite{snell2017prototypical}: exact L2-based nearest neighbor classification can then be performed in a single matrix-vector multiplication instead of requiring additional hardware for distance computation, a technique also used in Chameleon~\cite{blanken2025chameleonmatmulfreetemporalconvolutional}. This equivalent FC layer constitutes the scenario-specific head for FSL.

\subsubsection{CL Implementation in \abbr}

For CL, we largely follow the same procedure as for FSL.
In FSL, however, the equivalent FC layer can be computed instantaneously, as all support data is present at adaptation time.
In CL, on the other hand, we assume that new classes appear over time, denoted as $T\geq1$, after an initial adaptation phase, denoted as $T = 0$ (\autoref{fig:fig1}(c)).
Therefore, instead of calculating the FC parameters once and overwriting the previous ones, we append new entries to the PN's FC parameters over time for each additional class while keeping the previous FC parameters unchanged, thereby alleviating catastrophic forgetting.

\subsubsection{ZSL Implementation in \abbr}

To perform ZSL in \abbr, we use a relation network (RN)~\cite{sung2018learning}.
Although PNs also support ZSL~\cite{snell2017prototypical}, we choose an RN since it learns a distance function rather than assuming one (for example, L2 in PNs), which can improve performance for ZSL~\cite{sung2018learning}. 
Since ZSL uses two data modalities, we require two embedder NNs for the RN (\autoref{fig:fig1}(d)).
The first NN produces support embeddings on a per-class basis to learn from the provided semantic information. This NN can either be a TCN or a multi-layer perceptron~(MLP), depending on whether the semantic information is temporal or not. The second NN then produces the embedding for the to-be-classified sample in the target domain.
All embeddings are then concatenated and fed into an MLP, the so-called \textit{relation head}, for prediction. These last two operations constitute the scenario-specific head for ZSL.
In this way, the two embedder NNs specialize in producing high-quality embeddings in their input domain while the relation head learns to combine this information into a prediction.

\subsubsection{ICL Implementation in \abbr}

\looseness-100 For ICL, the dataflow is slightly different from the previous learning scenarios. 
Instead of concatenating support and query samples \emph{after} the embedding step like in ZSL, we concatenate all samples \emph{before} processing with the embedder network (\autoref{fig:fig1}(e)). For ZSL, this is not possible as the scenario incorporates two different data types and hence requires two different embedders. For ICL, however, this order reversal has two key implications.
First, it gives the embedder network the ability to directly model relationships between all inputs, instead of the relation head in ZSL. Second, it allows the TCN to produce one output prediction for each input, which is necessary for tasks such as regression or language modeling. 
After this joint-embedding step, a single next-output prediction can be calculated using one or more FC layers in an MLP, which forms the scenario-specific head for ICL.

\section{Results}
\label{sec:results}

This section presents the experimental results from using the \abbr~framework. First, \autoref{sec:hardware_eval} introduces our hardware test platform. Then, for each learning scenario (Sections~\ref{sec:results:fsl} to~\ref{sec:results:icl}), we describe the representative benchmark or dataset used in this work, followed by software results before presenting \abbr's performance when deployed in hardware. \autoref{sec:synopsis} finally outlines a synopsis of the obtained results.

\subsection{Test Platform}
\label{sec:hardware_eval}

\begin{figure}[t]
\centering
\includegraphics[width=0.48\textwidth]{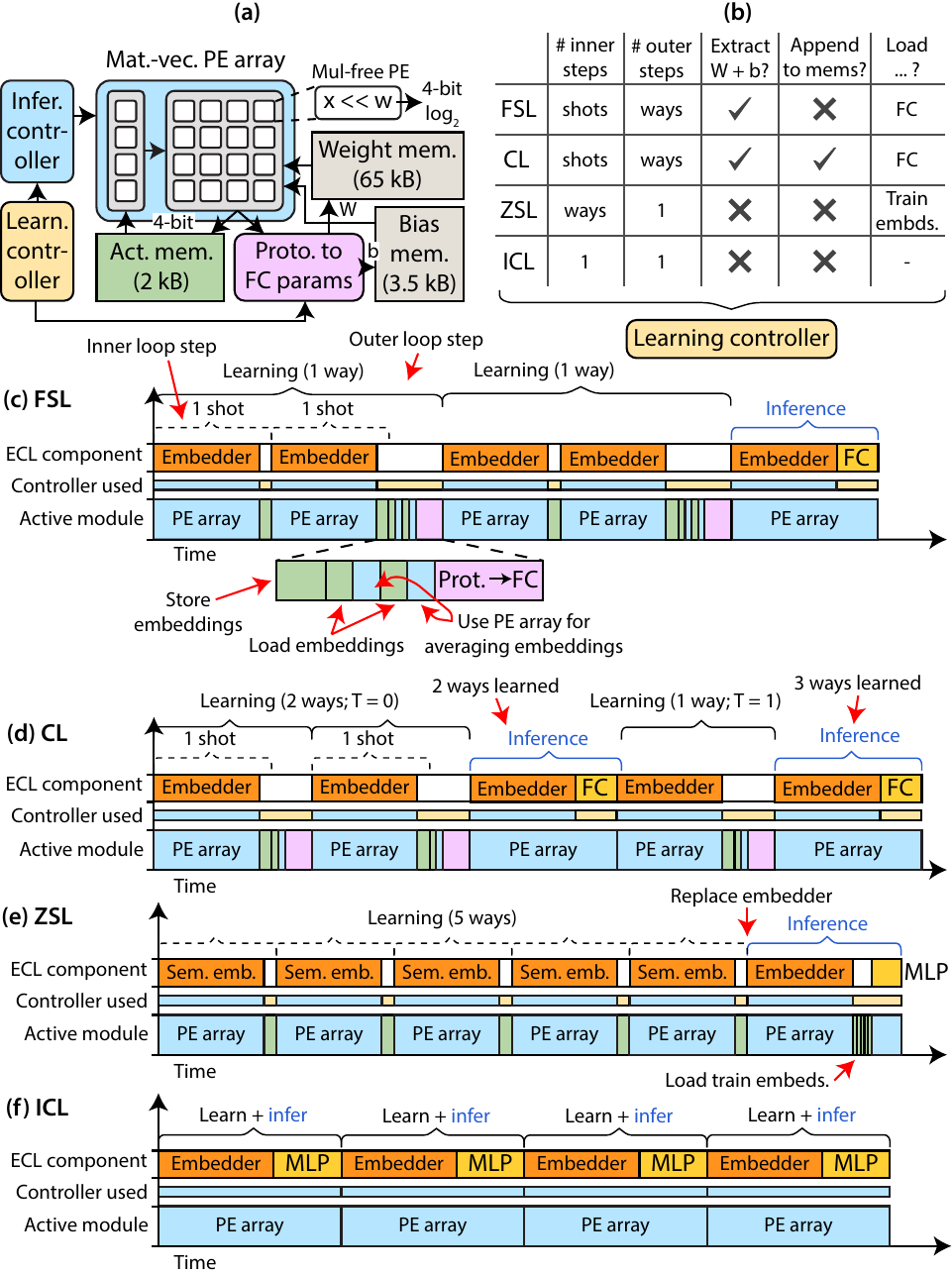}
\caption{\textbf{(a)} Simplified architecture overview of Chameleon. The SoC includes a matrix-vector processing element (PE) array which uses 4-bit $\log_2$ weights for multiplier-free PEs and 4-bit activations. Together with the inference control logic, it forms the SoC's NN inference data path. \textbf{(b)} Comparison of Chameleon's configurations per learning scenario, which are implemented by the SoC's learning controller. \textbf{(c)}-\textbf{(f)} Qualitative illustration of \abbr's execution schedules on Chameleon for FSL, CL, ZSL and ICL respectively. We show per scenario which main module and controller are used over time for the two ECL components. Overall, \abbr's mapping to the SoC can be partitioned in three phases: (i)~embedding all support samples for a single way (inner loop), (ii)~optionally converting them to prototypical parameters (outer loop), and (iii)~using the support embeddings, FC or MLP with the embedded query sample to predict. Dashed brackets indicate inner loop steps, while regular black brackets indicate outer loop steps. Blue brackets indicate the inference phase. Colors in (c)-(f) match the blocks in (a) and components in \autoref{fig:fig1}.}
\label{fig:timeline}
\end{figure}

To validate \abbr~in silicon, we use the Chameleon SoC~\cite{blanken2025chameleonmatmulfreetemporalconvolutional} as our test platform. Chameleon is an NN accelerator optimized for generating and processing dense embeddings from temporal data using TCNs. \autoref{fig:timeline}(a) outlines Chameleon's architecture. The SoC's matrix-vector processing element (PE) array uses inference-optimized quantization with 4-bit activations and 4-bit $\log_2$ weights, to shrink the memory footprint and reduce the multipliers to bit shifters. The NN weights and biases reside in 65~kB and 3.5~kB memories respectively while activations and embeddings share a 2~kB memory. An inference controller manages inference and embedding generation while a separate learning controller handles embedding processing for learning. \autoref{fig:timeline}(b) compares how the latter controller is configured to support each of \abbr's four learning scenarios.

These configurations result in distinct execution schedules. \autoref{fig:timeline}(c)-(f) show how \abbr's two key components map onto Chameleon over time for FSL, CL, ZSL, and ICL respectively.
Additionally, they indicate which main module and controller are used over time. 
All schedules execute \abbr~in three phases, as annotated in \autoref{fig:timeline}(c) for FSL.
First, the TCN embeds a support sample, and the SoC stores the resulting embedding in its activation memory. We call this an inner-loop step, which repeats a number of times dependent on the scenario (see~\autoref{fig:timeline}(b)). Second, if the scenario requires it, the stored support embeddings are converted into equivalent prototypical parameters and written to the same weight and bias memories that hold the embedder's parameters. Together, the first two phases form one outer-loop step, which also repeats a a number of times dependent on the scenario. Third, the query sample is embedded and passed to the scenario-specific head, \textit{i.e.}, one or a few FC layers, to make a prediction.
\autoref{fig:timeline}(d) shows these three phases for CL under \abbr~in a 1-shot use case. At $T=0$, a single-step inner loop and a two-step outer loop learn two ways, enabling inference over those two ways. A single inner and outer loop then learn an additional class, whose weights and bias are appended to the weight and bias memories, enabling inference over three ways.
In contrast, for ZSL (\autoref{fig:timeline}(e)), no parameter conversion is needed, reducing the outer loop to a single step but changing the inner loop step count to the number of ways. After embedding the semantic samples, the semantic embedder is replaced by the query-data embedder. Then, the SoC is ready for inference. For ICL (\autoref{fig:timeline}(f)), all samples are concatenated before processing by the embedder, collapsing both the inner and outer loops to a single step.
These schedules verify a key advantage of our embedder-centric formulation: extensive reuse of the inference data path for both TCN embedding generation and FC-layer computation.
As a result, \abbr's hardware efficiency hinges primarily on how efficiently inference is performed.
Since Chameleon is optimized for processing TCNs, it underpins the learning efficiency results that follow.

\subsection{\abbr~Supports On-the-fly Customization with FSL}
\label{sec:results:fsl}

We first demonstrate FSL under \abbr, where an NN can learn to recognize new classes from only a few labeled examples.

\subsubsection{Representative Benchmark}
To evaluate the performance of our framework on FSL for classification tasks, we use the popular Omniglot dataset~\cite{lake_omniglot}.
The dataset consists of a total of 1623 different handwritten characters across a large variety of alphabets, containing 20 sample images per character of $28\times28$ pixels. Since our framework is designed for processing sequential data, we flatten each input image to shape it into a 1D sequence.
We test on Omniglot across the standard evaluation settings of 1, 5-shot and 5, 20-way, as well as 32-way 1-shot.
We choose this dataset as it is widely used across both software and hardware works to measure FSL performance~\cite{maml_finn, snell2017prototypical, sapiens, fslhdnn}. 
Hence, Omniglot enables a direct comparison to these previous works.

\subsubsection{Software Results}

\looseness-100 \autoref{tab:pn_comp_with_cnn} shows the FSL accuracies across the standard FSL evaluation settings for both the TCN embedder that \abbr~employs and a commonly used convolutional neural network (CNN) alternative~\cite{snell2017prototypical, maml_finn}.
Compared to the similarly-sized CNN, the TCN incurs a drop of 0.2 to 1.6 accuracy points across all settings.
However, this is acceptable and expected, as neighboring pixels from the original image can be tens of timesteps apart when flattened for the TCN.
Hence, by using a sequential embedder in \abbr, we maintain support for learning from image data while we extend support for learning from variable-length sequences across modalities, which we will demonstrate in the remainder of this work.

\input{tables/pn_comp_with_cnn}
\begin{figure}[t]
\centering
\includegraphics[width=0.48\textwidth]{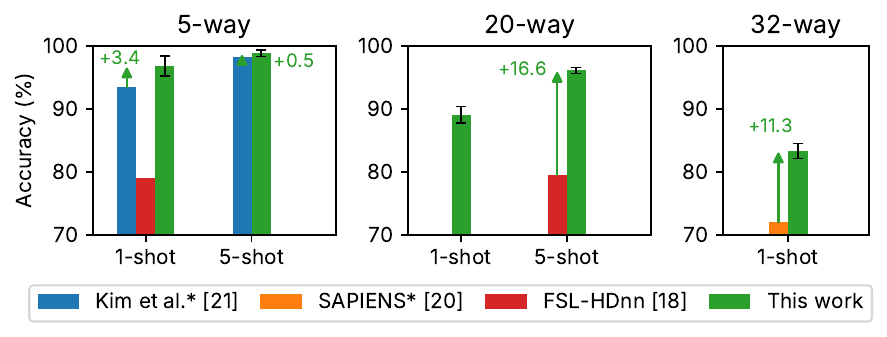}
\caption{FSL test accuracy comparison on the Omniglot dataset between FSL accelerators that have reported Omniglot results in silicon. We set new accuracy records across all shot-way scenarios, with improvement ranging from 0.5 to 16.6 accuracy points. Missing bars originate from non-reported accuracies. Error bars indicate 95\%-confidence intervals for our work. Green annotations indicate accuracy point improvements over the SotA. $^*$ indicates the use of an off-chip FP32 embedder.}
\label{fig:nmi_few_shot_comparison}
\end{figure}

\input{tables/e2e_omniglot}

\subsubsection{Hardware Demonstration}
\autoref{fig:nmi_few_shot_comparison} compares the FSL accuracies of the \abbr~framework on Chameleon to those of other silicon designs that perform FSL on Omniglot. These results for FSL were previously reported in~\cite{blanken2025chameleonmatmulfreetemporalconvolutional} and are included here to show the performance of \abbr~on a representative use case for FSL.
It can be seen that through our framework, we achieve new accuracy records across all shot-way scenarios versus prior works.
When comparing in detail with the FSL-HDnn chip~\cite{fslhdnn} (\autoref{tab:e2eomniglot}), which is, to the best of our knowledge, the only other end-to-end work with silicon results on Omniglot, we additionally demonstrate major gains in hardware efficiency.
We find that, at iso-clock-frequency, our framework reduces power by $2.3\times$, latency by $90\times$, and model size by $93\times$.
Furthermore, while FSL-HDnn~\cite{fslhdnn} requires external memory accesses for the NN, we eliminate these in our framework by specifically training a quantized NN that can be stored fully in on-chip SRAM.
In addition, the total silicon area overhead of the hardware required for learning with our framework is only 0.5\%, compared to 25\% for FSL-HDnn~\cite{fslhdnn}. This stems from FSL-HDnn's reliance on hyperdimensional computing (HDC)~\cite{kanerva2009hyperdimensional}, which is significantly more complex than our equivalent FC-layer forward pass and requires dedicated hardware.
By setting new accuracy records while significantly reducing the power, latency, and area, we validate \abbr's ability to efficiently support on-the-fly customization at the edge using FSL.

\subsection{\abbr~Supports Knowledge Accumulation with CL}
\label{sec:results:cl}

Second, we demonstrate CL under \abbr, where an NN can learn multiple tasks over time without overwriting knowledge of previously learned tasks, unlike with FSL.

\subsubsection{Representative Benchmark}

To validate the CL capabilities of our framework, we test it with the keyword FSCIL benchmark from the NeuroBench initiative~\cite{yik2025neurobench}.
The dataset consists of 100 spoken keyword classes for pre-training and 100 extra keyword classes for class-incremental CL, all with a duration of 1~s.
This high number of classes combined with the temporal sensory modality of the samples makes the keyword FSCIL benchmark an ideal fit for measuring the CL performance of \abbr.
During CL, we follow the original setting and learn the 100 keywords over ten 10-way 5-shot sessions.
To ensure a fair comparison with the baseline models from the original NeuroBench work~\cite{yik2025neurobench}, each \qty{48}{\kilo\hertz} audio sample is pre-processed off-chip into MFCCs.
By requiring a pre-trained model to learn new classes while retaining knowledge of previously learned ones, this benchmark aims to emulate a realistic use case at the edge.

\subsubsection{Software Results}

\begin{figure}[t]
\centering
\includegraphics[width=0.485\textwidth]{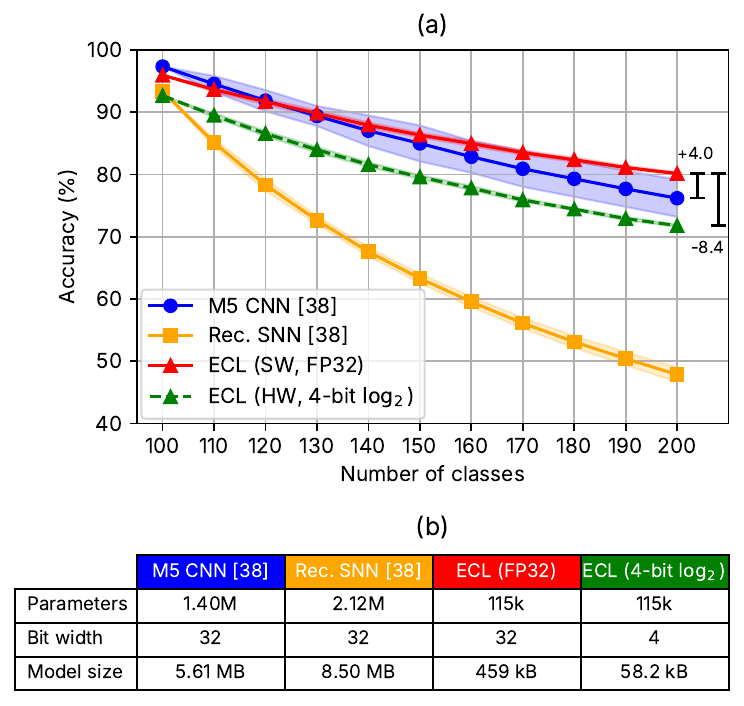}
\caption{\looseness-100 \textbf{(a)} CL accuracy comparison between \abbr~in software (SW) and in hardware (HW) on Chameleon and the NeuroBench baselines on the keyword FSCIL dataset~\cite{yik2025neurobench}. While \abbr~in FP outperforms both baselines after learning 20 new classes, our framework incurs a 4.4 accuracy-point drop compared to the $\sim96\times$ larger M5 baseline after continually learning the full dataset under tight quantization. Shaded areas indicate 95\%-confidence intervals. \textbf{(b)} Parameter count, bit width, and total model size comparison between the four models.}
\label{fig:mswc_cl_results}
\end{figure}

\looseness-100 \autoref{fig:mswc_cl_results} shows how the test accuracy varies with an increasing number of classes learned continually compared to two baseline models published in NeuroBench~\cite{yik2025neurobench}: the M5 CNN~\cite{m5} and a two-layer recurrent leaky integrate-and-fire (LIF) spiking neural network (SNN).
Compared to the CNN, \abbr's floating-point (FP) NN has a slightly lower accuracy from pre-training but starts to outperform it after learning only twenty classes with a final increase of four accuracy points, even though \abbr's NN is $\sim12\times$ smaller.
We postulate that our smaller embedder is forced by its capacity limit to learn more generic, reusable features, leading to a higher accuracy after the novel classes are introduced.
Furthermore, we find that our framework improves significantly over the SNN, even though \abbr's FP NN is $\sim18.4\times$ smaller.
Overall, our framework sets a new standard for CL on the keyword FSCIL dataset at a significant reduction in model size.

\subsubsection{Hardware Demonstration}

\looseness-100 \autoref{fig:mswc_cl_results} also displays the CL accuracies of the \abbr~framework when the model is quantized and executed on the Chameleon chip.
The $\sim8\times$ reduced model size of only 58.2~kB, compared to the FP \abbr~model, results in a drop of 8.4 accuracy points after learning 200 classes. %
Compared to the $96\times$ larger M5 model, our quantized model only incurs a loss of 4.4 accuracy points.
\autoref{tab:cl_metrics} reports the corresponding hardware performance of \abbr~for CL on this benchmark.
To the best of our knowledge, this is the first work to demonstrate end-to-end fully on-chip CL on the keyword FSCIL dataset from NeuroBench.
The only other end-to-end CL work with silicon results is Clo-HDnn~\cite{clo_hdnn}. However, it only supports learning up to 128 classes due to high dimensionality of the used embeddings (1024-8192). Additionally, similar to FSL-HDnn~\cite{fslhdnn}, it requires reloading the on-chip weights during inference as it cannot store all weights on-chip, introducing significant additional latency and energy penalties.
\abbr~addresses both limitations.
First, \abbr's much lower embedding dimension for CL (64) enables $4–32\times$ smaller embedding data sizes, so that 200 classes can be stored on-chip using only 6.4~kB.
Second, by compressing \abbr's embedding NN, we can store all model parameters fully on-chip.
Due to being fully end-to-end on-chip, \abbr~uses only \qty{9.5}{\micro\watt} on Chameleon for learning and inference while processing the keywords in real time.
By demonstrating the first end-to-end fully on-chip deployment on the NeuroBench keyword FSCIL dataset, we show our framework's ability to accurately learn 100 new classes with only a few samples and validate how \abbr~can be used to accumulate knowledge over time at the edge with CL.

\subsection{\abbr~Leverages Semantic Data with ZSL}
\label{sec:results:zsl}

Third, we demonstrate ZSL under \abbr, where an NN learns to recognize new classes from semantic information alone, \textit{i.e.}, without labeled data in the query domain, unlike FSL and CL.

\subsubsection{Representative Benchmark}

\input{tables/specs_cl}

To demonstrate ZSL with our framework, we use the Fluent Speech Commands (FSC) dataset~\cite{lugosch_speech_2019}. The FSC dataset consists of 248 unique spoken phrases up to 13~s long, each with a corresponding transcription. Every sentence concerns a single action, object, and location. For example: ``Bathroom (location) heat (object) down (action)''. We use 75\% of these phrases for training, 10\% for validation, and 15\% for testing, with no class overlap across splits. To enable ZSL on FSC, we use an MLP to create the semantic transcription embeddings, while a TCN embeds the test audio sample, which is converted to MFCCs beforehand. These embeddings are then combined and passed into the relation head to predict which never-\textit{heard}-before transcription was spoken. While FSC was originally proposed as a benchmark for spoken language understanding through supervised classification, we adopt it under a zero-shot setting.
In this setup, the semantic zero-shot data are the transcriptions of the to-be-classified sentences.
We opt for this benchmark as there are currently no datasets based on sequential data suitable for zero-shot learning in extreme-edge deployments.

\subsubsection{Software Results}

\looseness-100 Under a 5-way ZSL scenario on FSC, \abbr~achieves a test accuracy of 80.8\% with
FP precision. Since ZSL on FSC is introduced in this paper as a small-scale alternative to existing benchmarks, no direct comparisons currently
exist: Relation Network~\cite{sung2018learning}, for instance, demonstrated ZSL using
embedding networks ranging from approximately 10 to 50 million parameters to achieve 84.5\% test accuracy on a 10-way task. By
contrast, \abbr~uses a semantic embedder with only 120k parameters, a test
embedder with 96.7k parameters, and a relation head with 28.6k parameters, making
the total network approximately 40 to $200\times$ smaller than those in reference
works. 

\input{tables/specs_zsl}

\subsubsection{Hardware Demonstration}

\begin{figure}[t]
\centering
\includegraphics[width=0.5\textwidth]{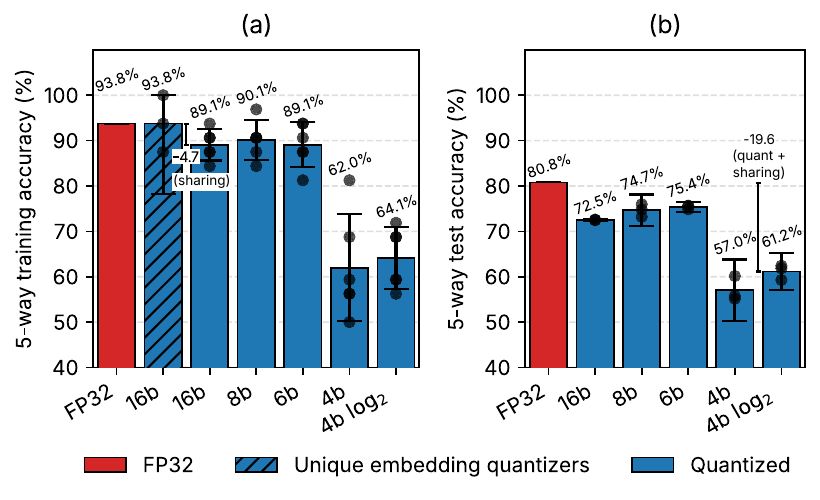}
\caption{Effect of weight and activation quantization on ZSL accuracy for FSC. \textbf{(a)}~Training accuracy at 2k steps. \textbf{(b)}~Test accuracy at the checkpoint with the highest validation accuracy. The FP32 bars show the baseline model that initializes all quantized runs. \textbf{(a)}~As Chameleon does not support per-channel scaling, both embedder NNs must share a single embedding quantizer. A unique quantizer per NN recovers the FP32 training accuracy at 16 bit, whereas sharing one costs 4.7 accuracy points, although the confidence intervals overlap. \textbf{(b)}~Test accuracy increases slightly from 16 to 6 bit but collapses at 4 bit. The 4-bit $\log_2$ weights supported by Chameleon perform better than standard 4-bit weights, but trail FP32 by 19.6 accuracy points, against 5.8 points for 6-bit quantization. 4-bit $\log_2$ accuracy here is higher than in \autoref{tab:specs_zsl} because this ablation uses a larger bias bit width than Chameleon supports. Hyperparameters are identical across runs. Bars show the mean across $n$ runs, points individual runs, and error bars 95\% confidence intervals.}
\label{fig:bitwidth_accuracy}
\end{figure}

\autoref{tab:specs_zsl} reports the performance of \abbr~for ZSL on the FSC dataset when deployed on Chameleon. To the best of our knowledge, this is the first demonstration of ZSL using semantic data at a \qty{}{\micro\watt} power budget suitable for the edge. The only other work that performs ZSL \textit{in silico} with a similar power budget is the work by Liu~\etal~\cite{liu2024high_zsl_eeg}. In that work, EEG data are used to adapt a seizure-prediction model to unseen patients, enabling learning without requiring seizure recordings from the new patient. However, their approach has three key limitations. First, it does not support semantic data, so it cannot generalize to truly unseen classes: all learned classes were seen during training. Second, it requires replay of data from previous patients, incurring energy-expensive off-chip memory accesses. Third, it relies on backpropagation for learning, requiring 16-bit data widths and incurring a 7\% area overhead.
Our proposed \abbr~framework addresses all of these shortcomings. By leveraging semantic data, \abbr~enables genuine zero-shot generalization to five previously unseen classes without any replay requirement. In addition, by using an RN as part of \abbr~for ZSL, we can use inference-level bit widths during learning. Combined, this yields a learning latency of 383~ms to learn five new classes for a total learning energy of \qty{3.1}{\micro\joule}. During inference, we process each sample within its duration to match the audio data rate, by setting the clock frequency to 5.8~kHz, which leads to a real-time inference power of \qty{8.0}{\micro\watt}. While advantageous for tight energy or power constraints, our low-bit-width RN does impact ZSL accuracy, going from 80.8\% to 60.6\% when quantized. 
Two limitations of Chameleon explain this gap: (i) the lack of per-channel scaling support requires both embedder NNs to share a single embedding quantizer~(\autoref{fig:bitwidth_accuracy}(a)), which leads to a 4.7-point training accuracy drop at 16-bit quantization, and (ii) while 6-bit quantization only loses 5.8 points compared to FP32, the 4-bit $\log_2$ weight format incurs a 19.6-point drop~(\autoref{fig:bitwidth_accuracy}(b)).
By demonstrating end-to-end ZSL under \abbr~using semantic data at a \qty{}{\micro\watt} budget for the first time, we enable learning even when labeled target-domain data are unavailable and add a level of versatility beyond what FSL and CL alone can provide.

\subsection{\abbr~Supports Adaptation Beyond Classification with ICL}
\label{sec:results:icl}

Fourth, we demonstrate ICL under \abbr, where an NN can learn to perform regression or predict the next token in a sequence, going beyond the previous learning scenarios focused only on classification.

\subsubsection{Representative Benchmark}

\looseness-100 To demonstrate ICL under \abbr, we use RegBench~\cite{akyurek2024context_regbench_icl}, a dataset designed to assess in-context language learning ability.
The RegBench dataset consists of a collection of problem instances, each represented as a sequence of examples.
All examples within a given instance are sampled from the same probabilistic language~\cite{akyurek2024context_regbench_icl}.
The goal in each example is to predict the next token based on the previous input.
Since, with increasing context length, more examples have been shown, prediction performance should increase.
In this work, we follow the standard setting of the benchmark, with a vocabulary size of twenty tokens and between ten and twenty examples of at most fifty tokens.
This benchmark dataset aims to be simple enough for analysis in small models while still capturing the core characteristics of ICL in large language models.
This makes RegBench a good fit for this work, as we aim to demonstrate our framework's capabilities at the edge, constrained by limited memory and compute.

\begin{figure}[t]
\centering
\includegraphics[width=0.46\textwidth]{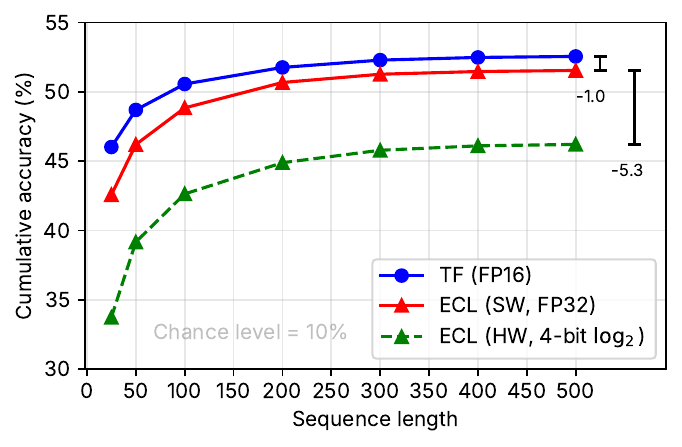}
\caption{\looseness-100 Cumulative ICL test accuracy comparison between \abbr~in software (SW) and in hardware (HW) on Chameleon and a parameter-matched TF on the RegBench dataset. Accuracy is measured over increasing sequence length for 5k query examples. While our FP model nears the transformer (TF) to within one accuracy point, when quantized, the final accuracy drops by 5.3 points.}
\label{fig:test_accuracy_regbench}
\end{figure}

\subsubsection{Software Results}
\autoref{fig:test_accuracy_regbench} shows how the cumulative test accuracy varies with increasing sequence length for both \abbr~at 133k parameters and a parameter-matched transformer (TF) baseline.
Despite \abbr's TCN model not having direct connections from all to all input tokens, unlike a TF~\cite{vaswani2017attention}, the TCN's FP accuracy approaches that of the TF within one accuracy point at the 500th token.
This result shows that, at a small scale, \abbr~achieves near-TF ICL performance while requiring 31$\times$ less activation memory storage (9.2 vs.\ 288~kB) by exploiting the TCN's dilation-induced sparsity~\cite{blanken2025chameleonmatmulfreetemporalconvolutional}.

\subsubsection{Hardware Demonstration}

\autoref{fig:test_accuracy_regbench} shows that the quantized model deployed on Chameleon reduces the cumulative accuracy by 
5.3 points compared to \abbr~in FP at the 500th token.
The quantized network occupies only 66.4~kB, making it suitable for execution on severely memory-constrained
edge devices.
We achieve a latency of 2.15~ms per token and
an average power consumption of 7.83~mW to yield an energy cost of 16.8~\si{\micro\joule} per token at a clock frequency of 100~MHz.
To the best of our knowledge, this is the first demonstration of ICL at a milliwatt power budget.
By showing how ICL can tackle this type of task, we demonstrate how \abbr~extends beyond plain classification, increasing the framework's versatility.

\subsection{Discussion}
\label{sec:synopsis}

\looseness-100 Putting together the results in Sections~\ref{sec:results:fsl} to~\ref{sec:results:icl}, we identify four common threads. First, while FSL is the only scenario that explicitly requires few labeled samples by definition,
we also demonstrate low-sample regimes for CL (5 samples per class), ZSL (0 samples in the target domain), and ICL (20-sample sequences).
Thus, across all four learning scenarios, \abbr~remains suitable for edge deployment, where labeled data are inherently scarce~\cite{blanken2025chameleonmatmulfreetemporalconvolutional}.
Second, by choosing the TCN as the embedder rather than modality-specific NN architectures, \abbr~operates across images (FSL on Omniglot), audio (CL and ZSL), and tokens (ICL on RegBench). To the best of our knowledge, no prior work has demonstrated learning across all three modalities on a single device at a micro-to-milliwatt power budget.
Third, since all learning scenarios share the same embedder NN type, and since all scenario-specific embedding processing logic can be incorporated into a single hardware block with only 0.5\% area overhead, \abbr~significantly expands accelerator versatility at minimal added cost.
Fourth, by supporting learning scenarios ranging from FSL to ICL, we support a progression of relaxing assumptions about available data at test time. Namely, FSL and CL assume labeled samples in the target modality while ZSL removes that assumption by substituting these with semantic data. ICL then removes the class structure entirely. This property allows \abbr~to be matched to the data available in a given use case.

Although above we presented the four learning scenarios as a progression, they are not necessarily mutually exclusive.
For example, combining FSL with CL yields few-shot class-incremental learning (FSCIL)~\cite{tao2020few_fscil}, where a model incrementally learns new classes with only a few shots per class. Likewise, combining ZSL with CL gives continual ZSL~\cite{chaudhry2018efficient, skorokhodov2020normalization,gautam2022tf} or lifelong ZSL~\cite{wei2020lifelong}.
More generally, ICL learns next-token prediction tasks, so it is compatible with FSL, CL, and ZSL once they are framed in this format.
One way to do this is to predict a class token given a context of the support embeddings and a query embedding.
However, each of these learning scenarios still requires an embedder network to embed all samples: adding an ICL mechanism on top increases the operation count, latency, and weight storage requirements, an overhead to be considered in light of the target deployment scenario specifications.

Taking a step back, we find that most prior works already rely on a fixed NN embedder for on-chip learning, yet none recognized this commonality explicitly or exploited it for hardware unification. Instead, prior works differentiated themselves by how the embeddings are used and which learning scenario is targeted. For example, SAPIENS~\cite{sapiens} and Kim~\etal~\cite{other_fsl_cim} apply L1 distance on top of embeddings for FSL, while FSL-HDnn~\cite{fslhdnn} and Clo-HDnn~\cite{clo_hdnn} use HDC~\cite{kanerva2009hyperdimensional} on embeddings for FSL and CL respectively. HDC expands embedding dimensionality to encode information, but this requires specialized hardware blocks~\cite{fslhdnn, clo_hdnn}; by contrast, \abbr~uses dense embeddings that map directly onto standard matrix-vector PE arrays, removing this overhead. Liu~\etal~\cite{liu2024high_zsl_eeg} take a different approach and apply BP to update the last two FC layers of a fixed CNN embedder for ZSL.
While each of these choices suit the target scenario well, they prevent the support of additional learning scenarios. Alternatively, Kwon~\etal~\cite{kwon2023tinytrain} selectively update parts of the full NN with BP, relaxing the fixed-embedder assumption entirely.
Yet, in this work too, this greater flexibility was not leveraged beyond FSL. The key insight of \abbr~is therefore not in any individual scenario implementation, but in recognizing this shared embedder-centric structure and exploiting it for hardware unification across four learning scenarios.

With regard to future work, we propose four key avenues.
First, one limitation of our evaluation is that we demonstrated \abbr~on relatively small datasets and NNs, reflecting the constraints of Chameleon as our extreme-edge demonstration platform.
However, the compute capabilities of edge devices are still increasing: they can then also deal with larger and more complex datasets.
Hence, future research should investigate how our framework scales, not only in terms of model and data size but also in task complexity.
Second, we postulate that \abbr~can be expanded to support simple few-shot reinforcement learning (RL) tasks by repurposing
\abbr's ICL mechanism, %
since RL problems can be viewed as sequence modeling problems that can also be learned in-context~\cite{chen2021decision_tf,janner2021offline_traj_tf}.
Third, \abbr~is currently formulated to be completely gradient-free for maximum hardware efficiency through simplicity.
Hence, a natural next step would be to explore whether allowing gradient descent steps, restricted to the scenario-specific heads, could further enhance performance while retaining hardware efficiency.
Fourth, weight-transport-free~\cite{frenkel_drtp} or forward-learning~\cite{hinton2022forwardforwardalgorithmpreliminaryinvestigations} methods could also be considered for the weight updates without incurring the cost of full backpropagation.
How such mechanisms can be efficiently incorporated into our unified embedder-centric framework is a promising avenue for future work.

\section{Conclusion}
\label{sec:conclusion}

In this work, we presented \abbr, a framework that unifies the FSL, CL, ZSL, and ICL online learning scenarios across sensory modalities.
We enabled this unification by framing every scenario in an embedder-centric way, which allowed close compatibility with existing inference hardware while keeping the embeddings to a few kB in memory.
To learn across sensory modalities, we used TCNs to support samples with long temporal dependencies.

To validate our framework's performance on a resource-constrained edge device, we used the Chameleon SoC.
In total, we considered four real-world use cases, one per learning scenario.
First, when performing FSL on the Omniglot dataset, we set new accuracy records across all shot-way scenarios, improving on the SotA by 0.5 to 16.6 accuracy points at $2.3\times$ lower power.
Second, using CL under \abbr, we provided the first end-to-end, fully on-chip hardware results on the keyword FSCIL task from NeuroBench. Compared to NeuroBench's original FP baseline, our model is $96\times$ smaller, incurring a loss of only 4.4 accuracy points after 200 classes at a real-time power of \SI{9.5}{\micro\watt}.
Third, we demonstrated the first end-to-end \textit{in-silico} implementation of ZSL using semantic data. In particular, we showed how \abbr~learned to classify five spoken sentences from the FSC dataset using only their transcription at an accuracy of 60.6\%. Learning consumes \SI{8.2}{\micro\watt}, while real-time inference afterward consumes only \SI{8.0}{\micro\watt}.
Fourth, we also demonstrated for the first time end-to-end, fully on-chip ICL at a milliwatt power budget. On the RegBench dataset, ICL under \abbr~requires 16.8~\si{\micro\joule} per token at a clock frequency of 100~MHz,
showing \abbr's ability to adapt beyond classification.

Together, these results establish \abbr~as a unified framework for versatile learning across scenarios at the edge, thereby enabling privacy-friendly, energy-efficient, and low-latency on-device adaptation.

\section*{Acknowledgement}

The authors thank Prof. Makinwa for his feedback, Dr. Marco P. Apolinario for his detailed input and for our fruitful discussions, and Dr. Johannes von Oswald for his RegBench implementation.

\printbibliography

\begin{IEEEbiography}[{\includegraphics[width=1in,height=1.3in,clip,keepaspectratio]{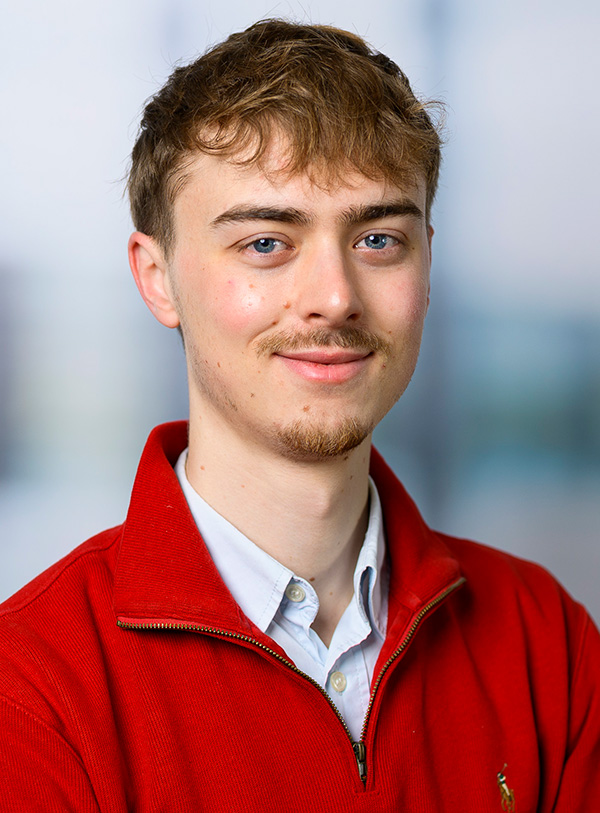}}]{Douwe den Blanken} (Graduate Student Member, IEEE) received the M.Sc. degree (\textit{with honors}) in embedded systems from Delft University of Technology (TU Delft), Delft, The Netherlands, in 2023, where he is currently pursuing the Ph.D. degree, under the supervision of Prof. C. Frenkel. His current research interests include efficient learning algorithms and their implementation in silicon, as well as the quantization and acceleration of modern DNNs. 
\end{IEEEbiography}

\begin{IEEEbiography}[{\includegraphics[width=1in,height=1.25in,clip,keepaspectratio]{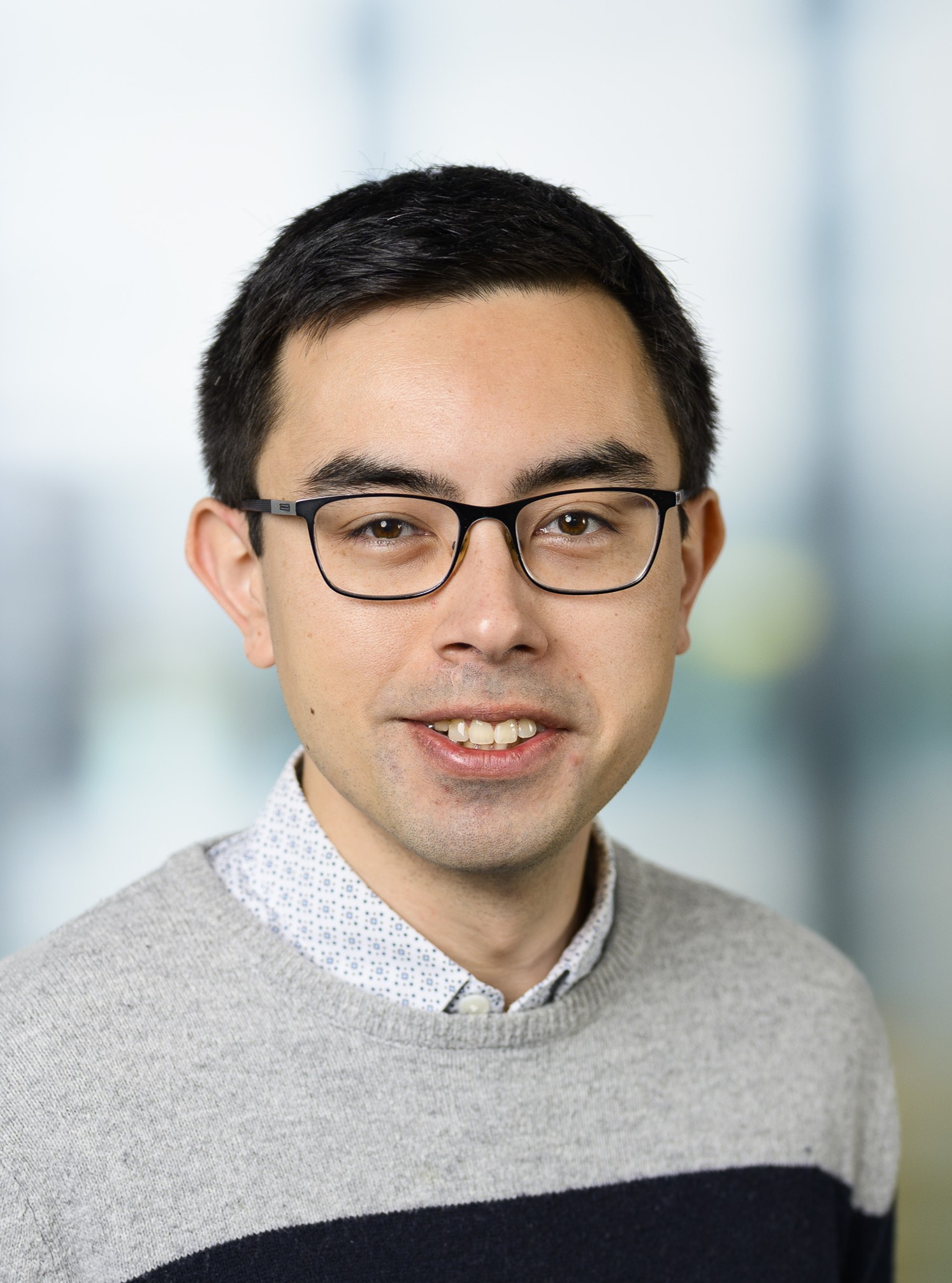}}]{Martin Lefebvre} (Member, IEEE) received the M.Sc. and Ph.D. degrees in engineering sciences from the Universit\'e catholique de Louvain (UCLouvain), Belgium, in 2017 and 2024.

His research interests include hardware-aware machine learning algorithms, mixed-signal vision chips for embedded image processing, and low-power current reference architectures. He currently is a post-doctoral researcher in the cognitive sensor nodes and systems (CogSys) laboratory led by Prof. Frenkel at TU Delft, The Netherlands, working on neuromorphic hardware/software co-design for efficient on-chip learning.

Dr. Lefebvre serves as a reviewer for various IEEE journals and conferences including IEEE Journal of Solid-State Circuits and IEEE Transactions on Circuits and Systems I and II.
\end{IEEEbiography}

\begin{IEEEbiography}[{\includegraphics[width=1.25in,height=1.3in,clip,keepaspectratio]{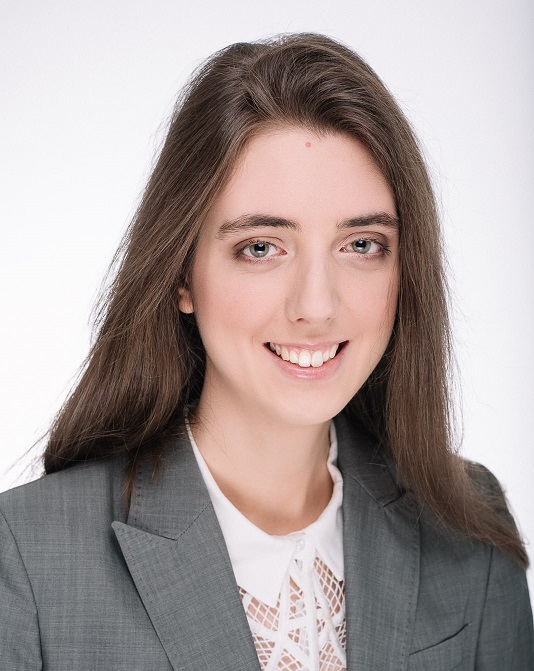}}]{Charlotte Frenkel}
(Member, IEEE) received the M.Sc. degree (\textit{summa cum laude}) in Electromechanical Engineering and the Ph.D. degree in Engineering Science from Universit\'e catholique de Louvain (UCLouvain), Louvain-la-Neuve, Belgium in 2015 and 2020, respectively. In February 2020, she joined the Institute of Neuroinformatics, UZH and ETH Zurich, Switzerland, as a postdoctoral researcher. She is an Assistant Professor at Delft University of Technology, Delft, The Netherlands, since July 2022, and a Research Scientist at Google since February 2026.

Her research aims at bridging the bottom-up (bio-inspired) and top-down (engineering-driven) design approaches toward neuromorphic intelligence, with a focus on hardware-algorithm co-design for (Neuro)AI, digital hardware accelerators, and brain-inspired on-device learning.

Dr. Frenkel received a best paper award at the IEEE International Symposium on Circuits and Systems (ISCAS) 2020 conference in the \textit{Neural Networks} track, and her Ph.D. thesis was awarded the FNRS-FWO / Nokia Bell Scientific Award 2021 and the FNRS-FWO / IBM Innovation Award 2021. In 2023, she was awarded prestigious Veni and AiNed Fellowship grants from the Dutch Research Council (NWO). She presented several invited talks, including keynotes at the tinyML EMEA technical forum 2021 and at the Neuro-Inspired Computational Elements (NICE) neuromorphic conference 2021. She serves or has served as a program co-chair of NICE 2023-2024 and of the tinyML Research Symposium 2024, as a TPC member of IEEE ISSCC for 2027 and IEEE ESSERC for 2022-2024, and as an associate editor for the IEEE Transactions on Biomedical Circuits and Systems for 2022-2025.
\end{IEEEbiography}

\vfill

\end{document}

%% file: tables/pn_comp_with_cnn.tex
\begin{table}[t]
\caption{Accuracy comparison between similarly sized FP32 TCN and CNN across standard evaluation settings for FSL on Omniglot using PNs.}
\label{tab:pn_comp_with_cnn}
\centering
\begin{tabular}{l c c cc cc}
\toprule
\textbf{Model} &
\multicolumn{2}{c}{\textbf{5-way}} &
\multicolumn{2}{c}{\textbf{20-way}} \\
& \textbf{1-shot} & \textbf{5-shot }&\textbf{ 1-shot} & \textbf{5-shot} \\
\midrule
CNN \cite{snell2017prototypical} & 98.8\% & 99.7\% & 96.0\% & 98.9\% \\
TCN & 97.9\% & 99.5\% & 94.4\% & 98.5\% \\
\bottomrule
\end{tabular}
\end{table}

%% file: tables/e2e_omniglot.tex
\begin{table}[t]
\caption{Comparison to FSL-HDnn~\cite{fslhdnn} on power, latency, model size and FSL area overhead. At iso-frequency, \abbr~is significantly more efficient, while relaxing latency constraints yields micro-watt level power. \textbf{Bold} indicates best.}
\label{tab:e2eomniglot}
\centering
\begin{adjustbox}{max width=0.48\textwidth}
\begin{tabular}{rccc}
\hline
 & FSL-HDnn~\cite{fslhdnn} & \multicolumn{2}{c}{This work} \\
\hline
Power & \qty{27}{\milli\watt} & \textbf{11.6~mW} & \textbf{12.9 \textmu W} \\
Latency & 53~ms & \textbf{0.59~ms} & 0.54~s \\
Frequency & 100~MHz & 100~MHz & 100~kHz \\
Model size & 5.5~MB & \multicolumn{2}{c}{\textbf{59~kB}} \\
All weights on-chip & \xmark & \multicolumn{2}{c}{\cmark} \\
FSL logic overhead & 25\% & \multicolumn{2}{c}{\textbf{0.5\%}} \\
Total core area & \SI{11.3}{\mm\squared} & \multicolumn{2}{c}{\textbf{\SI[detect-all]{0.83}{\mm\squared}}} \\
\hline
\end{tabular}
\end{adjustbox}
\end{table}

%% file: tables/specs_cl.tex
\begin{table}[t]
\caption{CL performance summary of \abbr~deployed on Chameleon for the NeuroBench keyword FSCIL dataset.}
\label{tab:cl_metrics}
\centering
\begin{threeparttable}
\begin{tabular}{r c}
\toprule
 & @ 0.73 V, \SI{14.4}{\kilo\hertz} \\
\midrule
Final accuracy (200 classes) & 71.8\% \\
Latency per sample & 1.0 s (real-time) \\
Average power & \SI{9.5}{\micro\watt} \\
Energy per sample & \SI{9.5}{\micro\joule} \\
Embedder network size & 58.2~kB \\
Total embeddings size (200 classes) & 6.4~kB \\
\bottomrule
\end{tabular}
\end{threeparttable}
\vspace{2mm}
\end{table}

%% file: tables/specs_zsl.tex
\begin{table}[t]
\caption{ZSL performance of \abbr~deployed on Chameleon for the FSC dataset. On this dataset, the latency and energy for inference are significantly higher than those for learning. This is due to the large size of the audio sample used for inference compared to the size of the semantic data used for learning.}
\label{tab:specs_zsl}
\centering
\begin{threeparttable}
\begin{tabular}{@{}rc@{}}
\toprule
 & @ 0.73V, 5.8 kHz\\
\midrule
Uses semantic data? & \cmark \\
5-way accuracy & 60.6\% \\
\hline
Learning power & \SI{8.2}{\micro\watt} \\
Total learning latency (5 ways) & 383 ms \\
Total learning energy (5 ways) & \SI{3.1}{\micro\joule} \\
\hline
Inference power & \SI{8.0}{\micro\watt} \\
Inference latency (per sample) & 13 s (real-time)\tnote{1} \\
Inference energy (per sample) & \SI{104}{\micro\joule} \\
\bottomrule
\end{tabular}
\begin{tablenotes}
  \item[1] Processing time matches maximum audio length (13~s); computation averages 16.4~ms/frame, matching one MFCC frame's duration. %
  \end{tablenotes}
\end{threeparttable}
\end{table}